\documentclass{Interspeech2024}
\usepackage{todonotes}
\usepackage{multirow}

\interspeechcameraready

\title{Children’s Speech Recognition through Discrete Token Enhancement}

\name{Vrunda N. Sukhadia, Shammur Absar Chowdhury}

\address{Qatar Computing Research Institute, HBKU, Qatar}
\email{sukhadiavrunda@gmail.com, shchowdhury@hbku.edu.qa}

\begin{document}
\maketitle

\begin{abstract}
Children's speech recognition is considered a low-resource task mainly due to the lack of publicly available data. There are several reasons for such data scarcity, including expensive data collection and annotation processes, and data privacy, among others. 
Transforming speech signals into discrete tokens that do not carry sensitive information but capture both linguistic and acoustic information could be a solution for privacy concerns.
In this study, we investigate the integration of discrete speech tokens into children's speech recognition systems as input without significantly degrading the ASR performance. Additionally, we explored single-view and multi-view strategies for creating these discrete labels. Furthermore, we tested the models for generalization capabilities with unseen domain and nativity dataset.
Results reveal that the discrete token ASR for children achieves nearly equivalent performance with an approximate 83\% reduction in parameters. 

\end{abstract}

\footnotetext{This paper was accepted at Interspeech 2024.}

\section{Introduction}

Automatic Children's speech recognition has recently attracted significant attention from research communities. One of the main reasons for such attention is that children increasingly interact with voice-activated assistants and technologies. This trend underscores the potential benefits of ASR technologies tailored for children, which can revolutionize learning tools, such as automated reading assessments \cite{Evanini2013} and interactive reading tutors \cite{Mostow2012} among others. These applications promise to enhance language acquisition for both native and non-native learners with immediate and multimodal feedback.

However, designing children's ASR has its unique challenges. Unlike adults, children's ASR is limited in resources and is still considered a low-resource task.
This is because there is a lack of large-scale publicly available children data, and collecting and annotating such datasets are expensive and also face many difficulties due to privacy and ethical considerations \cite{ClausEtAl2013,Wang2021LowResourceGermanASR,feng2024towards}. Moreover, many studies have consistently highlighted the disparities between child and adult ASR performance, especially in English, due to difficulties in acoustic and language modeling \cite{Lee1999AcousticsChildrenSpeech, Smith1992,KoenigLuceroPerlman2008,KoenigLucero2008,LeePotamianosNarayanan1999,LeePotamianosNarayanan1997,VorperianKent2007}.
The variabilities seen in children's speech data are due to the differences in speech development rates (inter-speaker variability) and evolving pronunciation skills within an individual child over time (intra-speaker variability). Moreover, children's speech includes significant mispronunciations and disfluencies, making it harder to annotate and model \cite{Yaruss1999LanguageDisfluency, Tran2020AnalysisDisfluency}.

Self-supervised learning (SSL) models have shown remarkable improvement in performance for various speech tasks \cite{baevski2020wav2vec,chen2022wavlm}, while reducing the dependency on extensively annotated datasets \cite{radford2023robust}. Studies such as \cite{FanAlwan2022,Fan2022DomainAdaptation,Jain2023Wav2Vec2ChildSpeech,Lahiri2023RobustEmbeddings,attia2023kid} have shown the efficacy of SSL models in improving child speech recognition, either using it for robust feature extractor or for finetuning the pre-trained model on specific datasets. Few studies have also been conducted to study the encoded information for children's speech present in the pre-trained SSL \cite{Shetty2023DevelopmentalFeatures, Lavechin2023BabySLM, YeungAlwan2018}.

Recent studies \cite{chang2023exploration,chang2023exploring2,ElKheir2024Beyond} have highlighted the usefulness of discrete speech units to represent speech signals, and their effects on ASR performance. Such compression not only reduces the storage and transmission size but also retains the essential acoustic and linguistic information while handling speaker variability better. This strategy also has the potential to handle privacy concerns, always faced when dealing with children's data. 

Therefore, in this study, we design an end-to-end English children's ASR system using discrete units as input to the models. Our proposed framework exploits the frame-level embeddings from pre-trained SSL models and quantizes them to a handful of discrete tokens considering representation either from a single SSL model (single view representation) or multiple (multi-view) SSL models using k-mean clustering models. These discrete tokens are then passed to an end-to-end ASR model.

We compare our proposed discrete ASR with an ASR trained on continuous embedding extracted from the pretrained HuBERT and WavLM model. 
Additionally, we compare the designed ASR system with results obtained using the state-of-the-art Whisper model \cite{radford2022robust} in both zero-shot and fine-tuned settings as the upper-bound for the study.
Furthermore, we show its efficacy when tested on unseen datasets, including (i) unseen domain, and (ii) non-native English datasets with both read and spontaneous speech style.

\noindent Therefore, our contribution in this paper includes:
\begin{itemize}
    \item Design and benchmark End-to-end Discrete ASR for children speech for native and non-native children datasets.
    \item Explore multi-view clustering strategy to design discrete tokens and compare it with the single-view method.
    \item Show the potential of the discrete children ASR for children ASR, while testing the generalization capability for the unseen domain, speaking styles, and nativity compared to the state-of-the-art Whisper model family.
\end{itemize}
\noindent To the best of our knowledge, this is the first study to explore the effectiveness of discrete tokens in single and multi-view settings for children ASR. 

\section{Methodology}
\begin{figure}[ht]

    \graphicspath{ {images/} }
    \scalebox{0.85}{ \includegraphics[width=8cm]{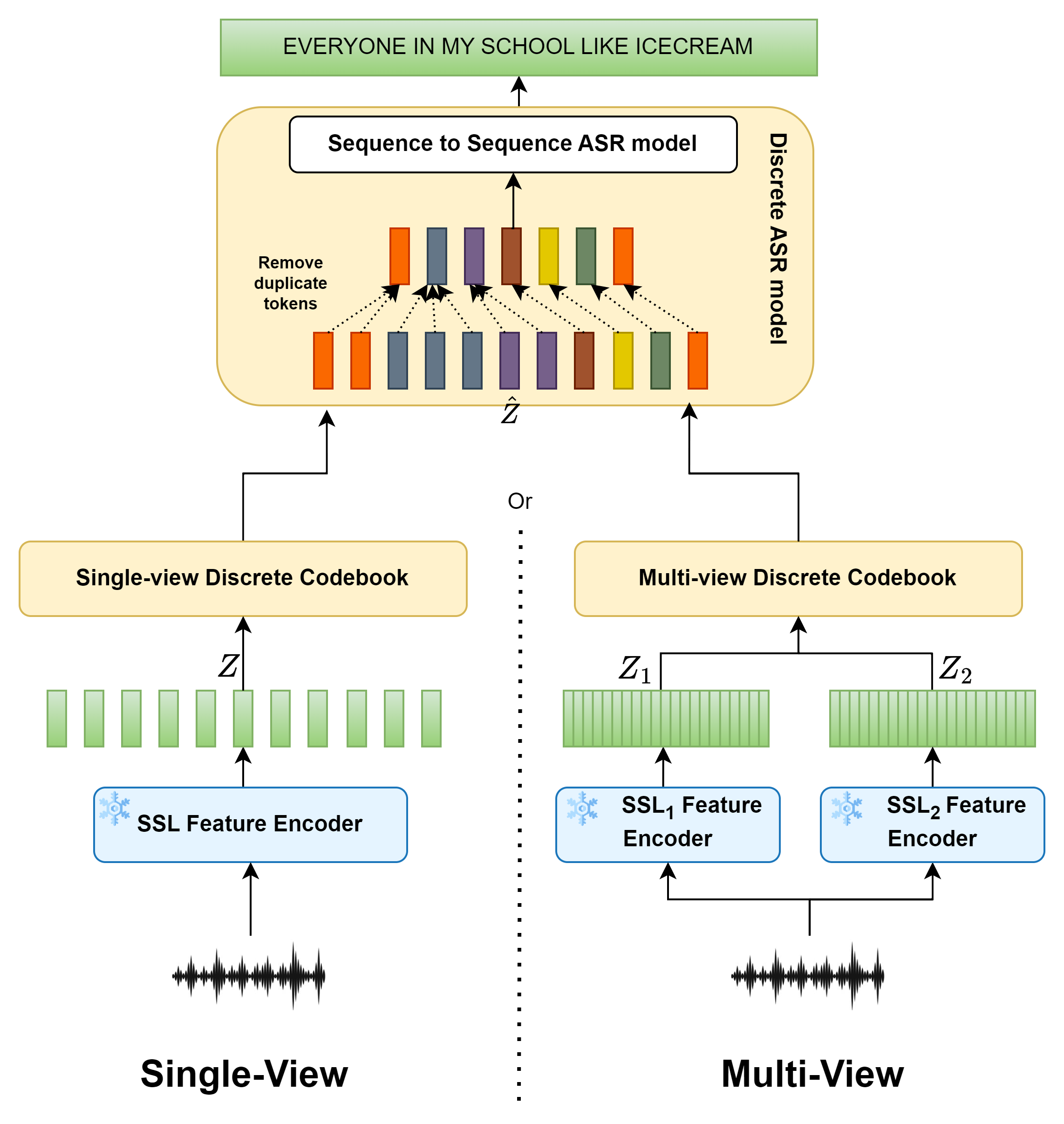}}
    \centering
    \caption{Discrete children ASR with single-view and multi-view discrete input.}
    \label{fig:exp_flow}
    \centering
\end{figure}

Figure \ref{fig:exp_flow} gives an overview of our proposed discrete Children ASR. Given an input utterance $X=\left [ x_1, x_2, \cdots , x_T \right ]$ of $T$ frames, the frame-level representation ($Z$) is first extracted from a \textit{SSL pretrained} model. A discrete codebook $\mathbb{C}$ is then trained with the frame-level $Z$ from the sampled utterances. For training the discrete codebook, we followed two different strategies utilizing either single representation or multi-view representation from pretrained models. 
We then utilize the trained $\mathbb{C}$ to infer $\hat{Z} = \mathbb{C}(Z)$, and use the discrete labels as an input to the encoder-decoder ASR model. 

\subsection{Discrete Codebook} 
We opt for a simple vector quantization \cite{makhoul1985vector, baevski2020wav2vec} technique for approximating frame-level embeddings through a fixed codebook size. We utilize a sequence of continuous feature vectors $Z = \{z_1, z_2, \ldots, z_T \}$ and then assign each $z_t$ to its nearest neighbor in the trained codebook, $\mathbb{C}$, with the code $Q_{i} \in \mathbb{C}$ assigned to the centroid $G_{i}$.  
The resultant discrete labels are quantized sequence $\hat{Z} = \{\hat{z}_1, \hat{z}_2, \ldots, \hat{z}_T\}$. 

To train the codebook, we opt for two different strategies: \textit{(i)} Single-View ($D^{(S)}$), and \textit{(ii)} Multi-View Codebook ($D^{(MV)}$). For the single-view strategy, we trained a simple k-means cluster model using representation from a pretrained SSL model. 
Whereas, for the multi-view, we considered the representations (or views $V^{(1)}$ and $V^{(2)}$) from two different SSL models and trained k-means clustering model. 
Given the conditional independence of $V^{(1)}$ and $V^{(2)}$, the strategy maximizes (M) the log-likelihood of each view, given the expected values for the hidden variables of the other view from the previous iteration and then calculate the expectation (E) for the hidden variables for the given view model parameters. Hence, optimizing for parameters with EM \cite{1410262} for both views. 
The optimization process is terminated when the improvement in log-likelihood is plateaued for a fixed number of iterations in each view. 
The final discrete label (during inference) is then assigned to the cluster that has the largest averaged posterior over both views.


The resultant discrete labels $\hat{Z}$ are temporarily aligned with the $Z$ and include repeated or commonly co-existing units. We followed steps such as de-duplication and subword modeling to reduce such redundancies, as proposed in \cite{chang2023exploring2}. For de-duplication, we merge the consecutive subsequences of identical tokens into a single token. Following, we transform the discrete sequence into meta-tokens sequence by using the Sentencepiece unigram model \cite{kudo2018subword}.


\subsection{Pretrained SSL} 
Given an input utterance, we extracted the representation using the following pretrained models:
\begin{itemize} [leftmargin=*]
    \item \textbf{facebook/\textit{HuBERT}-large-ll60k}: \textit{\textit{HuBERT}} identifies acoustic units by employing a clustering method to generate target labels corresponding to input features. Subsequently, masking is employed on the input features, and training is carried out to minimize the masked prediction loss using cluster labels as targets. 
    This model comprises \textbf{316M parameters}.
    \item \textbf{microsoft/\textit{WavLM}-large:} WavLM introduces gated relative position bias into the transformer architecture. In addition to employing masked prediction loss akin to \textit{\textit{HuBERT}}, it also integrates a denoising task during self-supervised learning. This model comprises \textbf{316M parameters}.
\end{itemize}
\subsection{ASR Architecture}
\label{architecture section}
For the children ASR, E-Branchformer \cite{kim2022ebranchformer} encoder and Transformer decoder \cite{vaswani2023attention} architecture is trained jointly with Connectionist Temporal Classification (CTC)/attention multi-task learning.
E-Branchformer is an improved version of Branchformer \cite{peng2022branchformer} with two parallel Macaron-style feed-forward network branches, with one branch responsible for capturing global context using multi-head attention, while the second branch captures local contextual information using multi-layer perceptron with convolutional gating (cgMLP). Following, the two branches are merged by concatenation operation, a 1-D depth-wise convolution, and a linear projection. The transformer decoder is used as the decoder part for the sequence-to-sequence model. The transformer decoder comprises an extra masked self-attention layer on top of an MHSA and a feed-forward layer.
The hyperparameters used for experiments are as shown in Table~\ref{hyperparameter table}.
\begin{table}[ht]
\centering
\vspace{-0.3cm}
\caption{Discrete Children ASR Model Configuration }
\scalebox{0.9}{
\begin{tabular}{@{}l|c@{}}
\textbf{Hyperparameters}                                                               & \textbf{Values} \\ \midrule
Kernel Size        & 31   \\ \hline
Feature dimension  & 512  \\ \hline
\# encoder layers  & 12   \\ \hline
Encoder units      & 1024 \\ \hline
\# decoder layers  & 6    \\ \hline
Decoder units      & 2048 \\ \hline
Attention heads    & 4    \\ \hline
\begin{tabular}[c]{@{}l@{}}Number of target \\ BPE (byte pair encoding)\end{tabular} & 5000            \\ \hline
\begin{tabular}[c]{@{}c@{}}Number of \\ source BPE\end{tabular}                        & 6000            \\ \hline
Number of clusters & 2000 \\ \hline
CTC weight         & 0.3 \\\hline
\end{tabular}}
\vspace{-0.2cm}
\label{hyperparameter table}
\end{table}



\section{Experimental Settings}
\subsection{Dataset}
\textbf{The My Science Tutor (\textit{MyST}) Corpus} \cite{pradhan2023science} is a collection of American English datasets featuring child speech, totaling over 393 hours from grades 3 to 5. The dataset features dialogs between the virtual tutors and the students, discussing various scientific concepts.
For the empirical study, we opt for 221 hours of the transcribed dataset, filtering out the very short (0.1 seconds and below) and too long (60 seconds and above) utterances. This preprocessing helped to reduce the computation memory needed to train the models. Following, we use the official data splits as train (167.48 hours), validate (25.60 hours), and test (27.95 hours) dataset. For discrete codebook training, 10\% of the training dataset, which amounts to 16.7 hours, is used.


\noindent \textbf{The CMU Kids Speech Corpus}\footnote{\url{http://www.ldc.upenn.edu/Catalog/LDC97S63.html}} is a collection of children's speech datasets containing 76 speakers, where the majority of the speakers are from grades 1 to 3. The age range of the children spans from six to eleven years old, with a distribution of 24 male and 52 female speakers.
The whole dataset includes a total of 5180 read utterances. We opted to use only $\approx$ 2.06 hours (22 \% of total data) of read-sentences as the unseen domain and age test set.\footnote{Utterances considered for the test have in-depth error analysis; the ids and information are collected from \url{https://isip.piconepress.com/projects/speech/databases/kids_speech} }

\noindent \textbf{Non-Native children's speech corpus} \cite{radha2022audio} is a collection of English read and spontaneous speech data from 20 bilingual (Telugu-English) children aged 8 to 12 with English proficiency. The dataset is gender-balanced (11 female and 9 male speakers) and is essential to test our proposed model's generalization capabilities for non-native speakers. 


    

\subsection{ASR Experiments}

\paragraph*{Baselines} We opt for two strong ASR baselines using the pretrained SSL models: HuBERT and WavLM. These models serve as feature extractors, providing rich and continuous contextual representations. The final input representation is obtained by computing the weighted sum of the embeddings from all layers. Following, we use the same encoder-decoder ASR architecture mentioned in Section \ref{architecture section}. These baseline models serve as reference points to measure the relative performance of the Discrete ASR model.

\paragraph*{Toplines} We compare the performances of the discrete ASRs with the readily available Whisper \cite{radford2022robust} models in Zero-shot and fine-tuned settings to understand the upper-bound performance. Whisper is a Transformer-based encoder-decoder model trained on 680,000 hours of labeled speech data annotated through weak supervision. These models underwent training using multilingual datasets. 
For the zero-shot settings, we present upper-bound results using two different model sizes:\footnote{\url{https://huggingface.co/openai/{whisper-small.en, whisper-medium.en}}} small (244M parameters) and medium (769M parameters). We utilize checkpoints for models trained exclusively on English data for the ASR task using 563,000 hours of data.
We fine-tuned the whisper model with 55 hours of MyST training set to mimic few-shot. Similarly, we also evaluate the Whisper model fine-tuned on the entire MyST training data, utilizing both\footnote{\url{https://huggingface.co/aadel4/{kid-whisper-small-en-myst, kid-whisper-medium-en-myst} }} the small and medium English-only model checkpoints.

\subsection{Model Training}
\subsubsection{Discrete Codebook Training}
The codebook responsible for generating discrete tokens from the SSL features is trained using k-means clustering in both single-view and multi-view scenarios. Consistent settings are applied to ensure a fair comparison between the two methods. For all the settings, the number of clusters is set to 2000, motivated by the success reported in \cite{chang2023exploration}, providing a fine granularity in representing the feature space. The k-means++ initialization method is used to enhance the clustering process.
Additionally, the number of random initializations ($N_{init}$) is set to 10, considering multiple starting points to achieve a better overall clustering solution. The maximum number of iterations ($max_{iter}$) is limited to 100.
These settings balance computational efficiency with clustering accuracy, effectively capturing the essential characteristics of the SSL features. 
By maintaining these consistent parameters across both single-view and multi-view scenarios, we aim to provide a robust comparison of the clustering performance and the resulting impact on the discrete token generation.

    
\subsubsection{ASR Model training}
The architecture specified in section \ref{architecture section} is adopted for training single-view and multi-view discrete ASR models. The \textit{ESPnet} \cite{watanabe2018espnet} recipe\footnote{\url{https://github.com/espnet/espnet/tree/master/egs2/librispeech_100/asr2}} is employed for training, utilizing two 32GB V100 GPUs. The models are trained using a learning rate of 0.002, with a warmup learning rate scheduler and the Adam optimizer across 100 epochs. Additionally, to augment the training data and enhance model robustness, the SpecAugment technique is applied to the input, facilitating better generalization.


% ...



\begin{table}[!ht]
\centering
\caption{Reported WER ($\downarrow$) presenting the baselines (\textit{HuBERT}-E2E and \textit{WavLM}-E2E) and the topline results using Whisper pre-trained model in zero-shot (0) and fine-tuned with 55 hours and all (All) \textit{MyST} training data. $\Delta$= $|WavLM-D^{(S)}-*|$, where $^*$ is different ASR results. Whisper-{S, M}: Whisper small (244M parameters) and medium (769M) models. Discrete token results are reported using \textit{HuBERT} and \textit{WavLM} models here.}
\scalebox{0.9}{
\begin{tabular}{lc}
\toprule
\multicolumn{1}{c}{\textbf{Models}} & \textbf{WER} ($\Delta$)  \\ %
\midrule
\multicolumn{2}{c}{Discrete Single-View ASRs}       \\ \midrule
\textit{HuBERT}-$D^{(S)}$                                & 15.65        \\
\textit{WavLM}-$D^{(S)}$                                 & 14.22      \\\midrule
\multicolumn{2}{c}{Baseline ASRs}               \\\midrule
\textit{HuBERT}-E2E                           & 14.98 (0.67) \\
\textit{WavLM}-E2E                           & 13.27 (0.95) \\\midrule
\multicolumn{2}{c}{Topline ASRs}                \\\midrule
Whisper-S (0)               & 13.93 (0.29) \\
Whisper-M (0)               & 12.9 (1.32)   \\ \hline
Whisper-S (55 hrs)                   & 13.23 (0.99) \\
Whisper-M (55 hrs)                   & 14.4 (0.18)   \\ \hline
Whisper-S (All)                     & 9.11 (5.11)  \\
Whisper-M (All)                     & 8.91 (5.31) \\
\bottomrule
\end{tabular}}
\vspace{-0.2cm}
\label{tab1:baseline}
\end{table}

\begin{table}[!ht]
\centering
\caption{Reported WER ($\downarrow$) presenting the results with discrete labels using \textit{HuBERT} and \textit{WavLM} for single- and multi-view representation along with the topline results using Whisper pre-trained model in zero-shot (0) and fine-tuned with 55 hours and All \textit{MyST} training data. U: Unseen domain/data. Whisper-M: Whisper medium model (769M parameters). All discrete models have 40.36M parameters. CMUk: CMU kids test subset.}
\scalebox{0.9}{
\begin{tabular}{lcccc}
\toprule
\multicolumn{1}{c}{\textbf{WER}}    & \textbf{Seen}               & \textbf{U: Domain} & \multicolumn{2}{c}{\textbf{U: Non-native}} \\ \midrule
\multicolumn{1}{c}{\textit{Models}} & \textit{MyST}               & \textit{CMUk}           & \textit{Read}         & \textit{Spont.}         \\ \midrule
\multicolumn{5}{c}{Single-View Discrete Tokens}                                                                                               \\\midrule
\textit{HuBERT}-$D^{(S)}$                                 & 15.65                        & 47.78                    & 38.40                  & 64.63                    \\
\textit{WavLM}-$D^{(S)}$                                 &  \textbf{14.22} & \textbf{45.60}           & \textbf{32.01}                  & \textbf{60.84}             \\\midrule
\multicolumn{5}{c}{Multi-View Discrete Tokens}                                                                                                \\\midrule
$D^{(MV)}$                                & 15.37                        & 46.60                    & 38.20                  & 63.35                    \\\midrule
\multicolumn{5}{c}{Topline}                                                                                                          \\\midrule
Whisper-M (0)                               & 12.9                        & 32.1                    & 30.38                 & 50.59                   \\
Whisper-M (All)                             & 8.91                        & 47.64                   & 37.71                 & 49.57                  \\ \bottomrule
\end{tabular}}
\vspace{-0.2cm}
\label{tab2:unseen}
\end{table}

\section{Results}
We reported the Word Error Rate (WER) for all the ASRs. The WER results are computed on normalized text, utilizing the \textit{BasicTextNormalizer} from Whisper \footnote{\url{https://github.com/openai/whisper/blob/main/whisper/normalizers/basic.py}}.

\subsection{Traditional {\em vs} Discrete Input}
Table \ref{tab1:baseline} reports WER for discrete token ASRs and compares it with the baselines and variations of Whisper - small and medium models in zero-shot, few (55 hours) shots, and fully fine-tuned settings. From the reported WER, we observed that discrete tokens perform comparably to the \textit{HuBERT}  and \textit{WavLM} end-to-end model, with a small performance drop of  $\Delta(WER)=0.67$ and $\Delta(WER)=0.95$ respectively. 
When compared with Whisper model variants (both zero- and few-shots), we noticed a maximum drop of $\Delta(WER)=1.32$. While with full \textit{MyST} training data, the drop goes to $\Delta(WER)=5.31$. All the aforementioned reported $\Delta(*)$ is w.r.t \textit{WavLM}-$D^{(S)}$. Considering the model sizes (Whisper Medium: 769M, Whisper Small: 242M, and Discrete Token ASR: 40.36M) and the extensive data utilized in Whisper's pre-training and subsequent fine-tuning, the performance of the Discrete Token ASR demonstrates nearly equivalent results while achieving an $\approx 83\%$ reduction in model size compared to Whisper Small and a $\approx 94\%$ reduction compared to Whisper Medium.

Moreover, discrete ASR efficiently reduces data sizes and input length as discussed above. For example, for a $T$ second utterance, the raw input signal (of 16 kHz sampling rate and 16-bit signed integer form) will need $16 \times 16000 \times T$ bits to encode; for SSL-based features with the rate of 50 frames per second, stored as float vectors and output embedding dimension of 1024 from one layer, we need $32 X 1024 X 50 X T$ bits. For discrete labels, we only need $11 X 50 X T$ bits for a maximum of 2048 clusters (11-bit) without even considering further improvement with de-duplication of sequence and subword modeling.

\subsection{Single-view {\em vs} Multi-view}
For the study, we exploit two simple ways to convert continuous speech features into discrete units. Using single-view and multi-view strategies, we reported the results on MyST in Table \ref{tab2:unseen}. We observed that in a single-view setup, the \textit{WavLM} discrete tokens outperform the HuBERT discrete tokens by 1.43 WER. 
We hypothesize that WavLM model embeddings are more robust due to its added utterance-mixing strategy, addressing the variability in child speech more efficiently.  

For multi-view setup, the performance of the $D^{(MV)}$ is superior to the HuBERT-$D^{(S)}$ model. However, WavLM-$D^{(S)}$ still outperforms both the variations. This potentially indicates that the selection of robust SSL models is essential to harness the power of multi-view discrete tokens. We keep this as a future exploration.

\subsection{Generalization Capabilities}
To test the generalization capabilities of these discrete token ASRs, we evaluated two unseen test sets and reported WER with single-, multi-view discrete ASRs along with the Whisper medium models in zero-shot and full (fine-tuned with full training data as the discrete models) settings in Table~\ref{tab2:unseen}.
We observed similar performance patterns across the datasets -- with different age groups (CMU kids data), nativity (non-native data), and speaking style (read- and spontaneous corpus). Similar to our previous observation, WavLM-$D^{(S)}$ outperforms all other discrete ASR systems and also gives comparable results to zero-shot Whisper models.


\begin{table}[ht]
\centering
\vspace{-0.3cm}
\caption{Example of Discrete ASR outputs}
\vspace{-0.2cm}
\scalebox{0.9}{
\begin{tabular}{l}
\toprule
\textbf{Ref:} A butterfly starts as an egg                                                                                                        \\\midrule
\begin{tabular}[c]{@{}l@{}} \textbf{ Verbatim: }{[}noise{]} a butterfly starts \textcolor{blue}{/EH/} {[}human\_noise{]} \\  an egg {[}human\_noise{]} {[}noise{]}\end{tabular} \\\midrule
\textbf{WavLM-$D^{(S)}$:} a butterfly starts \textcolor{red}{I} as an \textcolor{red}{X}                                                                                                             \\\midrule
\textbf{$D^{(MV)}$:} a butterfly starts \textcolor{red}{E} as an egg                      \\ \bottomrule                                                                                    
\end{tabular}}
\vspace{-0.25cm}
\label{eg}
\end{table}
\subsection{Error Analysis}
For the study, we briefly studied the effect of added noises on the model performance. Our initial exploration suggests that, with the different errors present in all the discrete ASRs, the multi-view discrete ASR is closer to the verbatim form of the transcription.
For example, as shown in Table \ref{eg}, the multi-view ASR can recognize the word ``egg'' correctly, even though in spoken form the word is followed by significant human noises. Moreover, the inserted char ``E'' is closer to the phonemic $EH$ sound that was actually in the speech. Such fine-grained prediction could help to detect mispronunciation and disfluencies present in the data more effectively.



\section{Conclusion}
This study presents the first benchmark for children's speech recognition with discrete tokens as input. From our exploration of discrete children ASR, we observed a comparable ASR performance with a significant reduction in model size and computational costs. Moreover, the discrete ASR provides additional data privacy required when dealing with sensitive speech data like children's speech. Our findings reflect the potential for multi-view discrete ASR, exploiting ensemble information encoded in separate SSL models. Further future research will involve studying how to enhance these discrete tokens with views extracted from different SSL models with different ASR architectures.

\bibliographystyle{IEEEtran}
\bibliography{mybib}

\begin{thebibliography}{10}
\providecommand{\url}[1]{#1}
\csname url@samestyle\endcsname
\providecommand{\newblock}{\relax}
\providecommand{\bibinfo}[2]{#2}
\providecommand{\BIBentrySTDinterwordspacing}{\spaceskip=0pt\relax}
\providecommand{\BIBentryALTinterwordstretchfactor}{4}
\providecommand{\BIBentryALTinterwordspacing}{\spaceskip=\fontdimen2\font plus
\BIBentryALTinterwordstretchfactor\fontdimen3\font minus \fontdimen4\font\relax}
\providecommand{\BIBforeignlanguage}[2]{{%
\expandafter\ifx\csname l@#1\endcsname\relax
\typeout{** WARNING: IEEEtran.bst: No hyphenation pattern has been}%
\typeout{** loaded for the language `#1'. Using the pattern for}%
\typeout{** the default language instead.}%
\else
\language=\csname l@#1\endcsname
\fi
#2}}
\providecommand{\BIBdecl}{\relax}
\BIBdecl

\bibitem{Evanini2013}
K.~Evanini and X.~Wang, ``Automated speech scoring for non-native middle school students with multiple task types,'' in \emph{Proceedings of the INTERSPEECH}, 2013.

\bibitem{Mostow2012}
J.~Mostow, ``Why and how our automated reading tutor listens,'' in \emph{Proceedings of the International Symposium on Automatic Detection of Errors in Pronunciation Training (ISADEPT)}, 2012.

\bibitem{ClausEtAl2013}
F.~Claus, H.~Gamboa~Rosales, R.~Petrick, H.-U. Hain, and R.~Hoffmann, ``A survey about databases of children's speech,'' in \emph{INTERSPEECH}, 2013.

\bibitem{Wang2021LowResourceGermanASR}
J.~Wang, Y.~Zhu, R.~Fan, W.~Chu, and A.~Alwan, ``Low resource german asr with untranscribed data spoken by non-native children- interspeech 2021 shared task spapl system,'' in \emph{Proc. Interspeech}, 2021.

\bibitem{feng2024towards}
S.~Feng, B.~M. Halpern, O.~Kudina, and O.~Scharenborg, ``Towards inclusive automatic speech recognition,'' \emph{Computer Speech \& Language}, vol.~84, 2024.

\bibitem{Lee1999AcousticsChildrenSpeech}
S.~Lee, A.~Potamianos, and S.~Narayanan, ``Acoustics of children's speech: Developmental changes of temporal and spectral parameters,'' \emph{J. Acoustical Soc. Amer.}, 1999.

\bibitem{Smith1992}
B.~L. Smith, ``Relationships between duration and temporal variability in children's speech,'' \emph{The Journal of the Acoustical Society of America}, 1992.

\bibitem{KoenigLuceroPerlman2008}
L.~L. Koenig, J.~C. Lucero, and E.~Perlman, ``Speech production variability in fricatives of children and adults: Results of functional data analysis,'' \emph{The Journal of the Acoustical Society of America}, 2008.

\bibitem{KoenigLucero2008}
L.~L. Koenig and J.~C. Lucero, ``Stop consonant voicing and intraoral pressure contours in women and children,'' \emph{The Journal of the Acoustical Society of America}, 2008.

\bibitem{LeePotamianosNarayanan1999}
S.~Lee, A.~Potamianos, and S.~Narayanan, ``Acoustics of children's speech: Developmental changes of temporal and spectral parameters,'' \emph{The Journal of the Acoustical Society of America}, 1999.

\bibitem{LeePotamianosNarayanan1997}
------, ``Analysis of children's speech: Duration, pitch and formants,'' in \emph{Fifth European Conference on Speech Communication and Technology}, 1997.

\bibitem{VorperianKent2007}
H.~K. Vorperian and R.~D. Kent, ``Vowel acoustic space development in children: A synthesis of acoustic and anatomic data,'' 2007.

\bibitem{Yaruss1999LanguageDisfluency}
J.~S. Yaruss, R.~M. Newman, and T.~Flora, ``Language and disfluency in nonstuttering children's conversational speech,'' \emph{J. Fluency Disord.}, 1999.

\bibitem{Tran2020AnalysisDisfluency}
T.~Tran, M.~Tinkler, G.~Yeung, A.~Alwan, and M.~Ostendorf, ``Analysis of disfluency in children's speech,'' in \emph{Proc. Interspeech}, 2020.

\bibitem{baevski2020wav2vec}
A.~Baevski, Y.~Zhou, A.~Mohamed, and M.~Auli, ``wav2vec 2.0: A framework for self-supervised learning of speech representations,'' \emph{Advances in neural information processing systems}, 2020.

\bibitem{chen2022wavlm}
S.~Chen, C.~Wang, Z.~Chen, Y.~Wu, S.~Liu, Z.~Chen, J.~Li, N.~Kanda, T.~Yoshioka, X.~Xiao \emph{et~al.}, ``Wavlm: Large-scale self-supervised pre-training for full stack speech processing,'' \emph{IEEE Journal of Selected Topics in Signal Processing}, 2022.

\bibitem{radford2023robust}
A.~Radford, J.~W. Kim, T.~Xu, G.~Brockman, C.~McLeavey, and I.~Sutskever, ``Robust speech recognition via large-scale weak supervision,'' in \emph{International Conference on Machine Learning}.\hskip 1em plus 0.5em minus 0.4em\relax PMLR, 2023.

\bibitem{FanAlwan2022}
R.~Fan and A.~Alwan, ``Draft: A novel framework to reduce domain shifting in self-supervised learning and its application to children's asr,'' in \emph{Proc. Interspeech 2022}, 2022.

\bibitem{Fan2022DomainAdaptation}
R.~Fan, Y.~Zhu, J.~Wang, and A.~Alwan, ``Towards better domain adaptation for self-supervised models: A case study of child asr,'' \emph{IEEE Journal of Selected Topics in Signal Processing}, vol.~16, no.~6, 2022.

\bibitem{Jain2023Wav2Vec2ChildSpeech}
R.~Jain, A.~Barcovschi, M.~Y. Yiwere, D.~Bigioi, P.~Corcoran, and H.~Cucu, ``A wav2vec2-based experimental study on self-supervised learning methods to improve child speech recognition,'' \emph{IEEE Access}, 2023.

\bibitem{Lahiri2023RobustEmbeddings}
R.~Lahiri, T.~Feng, R.~Hebbar, C.~Lord, S.~H. Kim, and S.~Narayanan, ``Robust self supervised speech embeddings for child-adult classification in interactions involving children with autism,'' in \emph{Proc. INTERSPEECH 2023}, 2023.

\bibitem{attia2023kid}
A.~A. Attia, J.~Liu, W.~Ai, D.~Demszky, and C.~Espy-Wilson, ``Kid-whisper: Towards bridging the performance gap in automatic speech recognition for children vs. adults,'' \emph{arXiv preprint arXiv:2309.07927}, 2023.

\bibitem{Shetty2023DevelopmentalFeatures}
V.~M. Shetty, S.~M. Lulich, and A.~Alwan, ``Developmental articulatory and acoustic features for six to ten year old children,'' in \emph{Proc. INTERSPEECH 2023}, 2023.

\bibitem{Lavechin2023BabySLM}
M.~Lavechin, Y.~Sy, H.~Titeux, M.~A.~C. Blandón, O.~Räsänen, H.~Bredin, E.~Dupoux, and A.~Cristia, ``Babyslm: language-acquisition-friendly benchmark of self-supervised spoken language models,'' in \emph{Proc. INTERSPEECH 2023}, 2023.

\bibitem{YeungAlwan2018}
G.~Yeung and A.~Alwan, ``On the difficulties of automatic speech recognition for kindergarten-aged children,'' in \emph{Interspeech 2018}, 2018.

\bibitem{chang2023exploration}
X.~Chang, B.~Yan, Y.~Fujita, T.~Maekaku, and S.~Watanabe, ``Exploration of efficient end-to-end asr using discretized input from self-supervised learning,'' \emph{arXiv preprint arXiv:2305.18108}, 2023.

\bibitem{chang2023exploring2}
X.~Chang, B.~Yan, K.~Choi, J.~Jung, Y.~Lu, S.~Maiti, R.~Sharma, J.~Shi, J.~Tian, S.~Watanabe \emph{et~al.}, ``Exploring speech recognition, translation, and understanding with discrete speech units: A comparative study,'' \emph{arXiv preprint arXiv:2309.15800}, 2023.

\bibitem{ElKheir2024Beyond}
Y.~E. Kheir, H.~Mubarak, A.~Ali, and S.~A. Chowdhury, ``Beyond orthography: Automatic recovery of short vowels and dialectal sounds in arabic,'' in \emph{Proceedings of the 62nd Annual Meeting of the Association for Computational Linguistics (ACL 2024)}, 2024.

\bibitem{radford2022robust}
A.~Radford, J.~W. Kim, T.~Xu, G.~Brockman, C.~McLeavey, and I.~Sutskever, ``Robust speech recognition via large-scale weak supervision,'' 2022.

\bibitem{makhoul1985vector}
J.~Makhoul, S.~Roucos, and H.~Gish, ``Vector quantization in speech coding,'' \emph{Proceedings of the IEEE}, 1985.

\bibitem{1410262}
S.~Bickel and T.~Scheffer, ``Multi-view clustering,'' in \emph{Fourth IEEE International Conference on Data Mining (ICDM'04)}, 2004.

\bibitem{kudo2018subword}
T.~Kudo, ``Subword regularization: Improving neural network translation models with multiple subword candidates,'' in \emph{Proceedings of the 56th Annual Meeting of the Association for Computational Linguistics (ACL)}, 2018.

\bibitem{kim2022ebranchformer}
K.~Kim, F.~Wu, Y.~Peng, J.~Pan, P.~Sridhar, K.~J. Han, and S.~Watanabe, ``E-branchformer: Branchformer with enhanced merging for speech recognition,'' 2022.

\bibitem{vaswani2023attention}
A.~Vaswani, N.~Shazeer, N.~Parmar, J.~Uszkoreit, L.~Jones, A.~N. Gomez, L.~Kaiser, and I.~Polosukhin, ``Attention is all you need,'' 2023.

\bibitem{peng2022branchformer}
Y.~Peng, S.~Dalmia, I.~Lane, and S.~Watanabe, ``Branchformer: Parallel mlp-attention architectures to capture local and global context for speech recognition and understanding,'' 2022.

\bibitem{pradhan2023science}
S.~S. Pradhan, R.~A. Cole, and W.~H. Ward, ``My science tutor (myst) -- a large corpus of children's conversational speech,'' 2023.

\bibitem{radha2022audio}
K.~Radha and M.~Bansal, ``Audio augmentation for non-native children’s speech recognition through discriminative learning,'' \emph{Entropy}, 2022.

\bibitem{watanabe2018espnet}
S.~Watanabe, T.~Hori, S.~Karita, T.~Hayashi, J.~Nishitoba, Y.~Unno, N.~{Enrique Yalta Soplin}, J.~Heymann, M.~Wiesner, N.~Chen, A.~Renduchintala, and T.~Ochiai, ``{ESPnet}: End-to-end speech processing toolkit,'' in \emph{Proceedings of Interspeech}, 2018.

\end{thebibliography}

\end{document}